\documentclass{article}

\usepackage{PRIMEarxiv}

\usepackage[utf8]{inputenc} 
\usepackage[T1]{fontenc}    
\usepackage{hyperref}       
\usepackage{url}            
\usepackage{booktabs}       
\usepackage{amsfonts}       
\usepackage{nicefrac}       
\usepackage{microtype}      
\usepackage{lipsum}
\usepackage{fancyhdr}       
\usepackage{graphicx}       
\graphicspath{{imgs/}}     
\usepackage{amsmath,amssymb,amsfonts}
\usepackage{cite, algorithmic, textcomp}
\usepackage{arydshln}
\usepackage{multirow}

\title{Synthetic Tabular Data Validation: A Divergence-Based Approach}

\author{
  Patricia A. Apellániz\thanks{\textbf{These authors contributed equally.}}\\
  Information Processing \\and Telecommunications Center \\
  ETS Ingenieros de Telecomunicación \\
  Universidad Politécnica de Madrid \\
  Madrid\\
  \textit{patricia.alonsod@upm.es} \\
  \And
  Ana Jim\'enez{\textsc {$^*$}}\\
  Information Processing \\and Telecommunications Center \\
  ETS Ingenieros de Telecomunicación \\
  Universidad Politécnica de Madrid \\
  Madrid\\
  \textit{ana.jimenezb@upm.es} \\
  \And
  Borja Arroyo Galende\\
  Information Processing \\and Telecommunications Center \\
  ETS Ingenieros de Telecomunicación \\
  Universidad Politécnica de Madrid \\
  Madrid\\
  \textit{borja.arroyog@upm.es} \\
  \And
  Juan Parras\\
  Information Processing \\and Telecommunications Center \\
  ETS Ingenieros de Telecomunicación \\
  Universidad Politécnica de Madrid \\
  Madrid\\
  \textit{j.parras@upm.es} \\
  \And
  Santiago Zazo\\
  Information Processing \\and Telecommunications Center \\
  ETS Ingenieros de Telecomunicación \\
  Universidad Politécnica de Madrid \\
  Madrid\\
  \textit{santiago.zazo@upm.es} \\
}

\begin{document}
\maketitle
\thispagestyle{fancy}

\begin{abstract}
The ever-increasing use of generative models in various fields where tabular data is used highlights the need for robust and standardized validation metrics to assess the similarity between real and synthetic data. Current methods lack a unified framework and rely on diverse and often inconclusive statistical measures. Divergences, which quantify discrepancies between data distributions, offer a promising avenue for validation. However, traditional approaches calculate divergences independently for each feature due to the complexity of joint distribution modeling. This paper addresses this challenge by proposing a novel approach that uses divergence estimation to overcome the limitations of marginal comparisons. Our core contribution lies in applying a divergence estimator to build a validation metric considering the joint distribution of real and synthetic data. We leverage a probabilistic classifier to approximate the density ratio between datasets, allowing the capture of complex relations. We specifically calculate two divergences: the well-known Kullback-Leibler (KL) divergence and the Jensen-Shannon (JS) divergence. KL divergence offers an established use in the field, while JS divergence is symmetric and bounded, providing a reliable metric. The efficacy of this approach is demonstrated through a series of experiments with varying distribution complexities. The initial phase involves comparing estimated divergences with analytical solutions for simple distributions, setting a benchmark for accuracy. Finally, we validate our method on a real-world dataset and its corresponding synthetic counterpart, showcasing its effectiveness in practical applications. This research offers a significant contribution with applicability beyond tabular data and the potential to improve synthetic data validation in various fields.\end{abstract}


\thispagestyle{fancy}

\section{Introduction}
\label{sec:introduction}
Generative models (GMs) are emerging as a disruptive force within artificial intelligence. By learning the underlying patterns and distributions of real-world data, these models can create novel synthetic data replicating the characteristics of the original domain. These synthetic data offer significant advantages for various research and development applications. In particular, synthetic data can be used for data augmentation, addressing limitations in data collection efforts. In addition, it can ensure privacy protection by replacing sensitive real-world information. Furthermore, it facilitates the exploration of rare or extreme scenarios (edge cases) that might be difficult or expensive to encounter in real-world settings. As a result, the synthetic data generated by these models is accelerating innovation in various fields, including healthcare\cite{Jadon2023LeveragingGA}, finance\cite{Caliskan2023ACA}, and the development of autonomous vehicles\cite{autonomousVehicles}, to mention some.

Following the impressive advances in GMs for data generation, a critical aspect emerges: ensuring the quality and effectiveness of synthetic data. Replicating the source data's statistical properties, accuracy, consistency, and domain-specific relevance is essential. This challenge becomes significantly pronounced as GMs gain traction within engineering design communities. The need for robust and standardized validation metrics becomes paramount. The current literature highlights a significant gap in this area. Although there are standardized evaluation metrics for the generation of synthetic images \cite{salimans2016improved}, \cite{heusel2018gans}, and the generation of text \cite{zhang2020bertscore}, \cite{10.3115/1073083.1073135}, measuring the quality of synthetic tabular data presents unique challenges. Unlike image data, qualitative evaluation through visual inspection is not feasible. Additionally, relying solely on expert insight can be highly inefficient. The existing landscape for tabular data validation often focuses primarily on the efficacy or utility in machine learning tasks \cite{che2017boosting}, \cite{8983215}. However, a consistent approach to assessing the similarity between synthetic and real tabular data remains sparse. Studies employ a diverse set of metrics, including pairwise correlation difference\cite{10.1007/978-3-030-59137-3_4}, support coverage\cite{gonsalves}, likelihood fitness\cite{xu2019modeling}, alongside various statistic values\cite{10.1007/978-3-030-59137-3_34},\cite{snoke2017general}. Reviews such as \cite{vicomtech} attempt to categorize these techniques, but a genuinely standardized approach or a single metric that captures the full spectrum of statistical information from a distribution remains a challenge.

This paper aims to address this gap by proposing a divergence estimator. Divergences offer a comprehensive approach to the validation of similarity. These measures quantify the discrepancy between two probability distributions, effectively capturing the differences between real and synthetic data. Unlike traditional methods that often focus on individual attributes \cite{hu2021tableganmca}, \cite{10.1007/978-3-030-61609-0_46}, divergences can consider the joint distribution of all attributes, providing a more holistic view of the data. However, it is essential to acknowledge the challenge associated with joint distributions. Modeling the intricate relations between all features in a dataset can be computationally expensive and intractable. Although some existing papers explore divergences or distances for validation, they often apply them to single attributes because of this limitation \cite{gonsalves}, \cite{zhao2021ctabgan}, \cite{tucker2020generating}. Crucially, even if individual features appear similar, the joint distribution, which captures their relations, may differ significantly. Estimating divergences between marginal distributions neglects the correlation between features. For example, in survival analysis databases, the risk associated with a disease is often closely related to the patient's age \cite{liu2019effect}. Although these two variables might be generated to mimic the original distributions, this method might fail to preserve their intrinsic, possibly nonlinear, correlation. In practice, disrupting original correlations can result in synthetic data that, while statistically similar in univariate terms, diverges significantly in multivariate contexts. This discrepancy is critical because it can affect the utility and reliability of synthetic data in applications where feature interdependencies play a crucial role, such as predictive modeling in healthcare. To address this issue, correlation matrices are sometimes employed to assess relations between features \cite{mckeever2020synthesising}. However, these matrices primarily reveal linear relations, provide complex representations usually interpreted graphically, and are challenging to analyze quantitatively. This underscores the importance of considering the joint distribution for an accurate assessment. Our proposed divergence estimator addresses this challenge by efficiently capturing these complex joint distributions. This approach enables the computation of divergences between real and synthetic data, allowing for a more comprehensive and robust evaluation of the quality of synthetic data. The advantages of using divergences for validation remain the same:
\begin{itemize}
    \item Sensitivity to Complex Distributions: Divergences can capture the full complexity of real-world data distributions, including nonlinear relations and rare events, which traditional methods might miss.
    \item Robustness to Noise: Divergences are less sensitive to noise and outliers, making them more reliable for practical applications.
    \item Interpretability: Divergences can be easily interpreted and visualized, providing valuable insights into the discrepancies between real and synthetic data. They condense statistical similarity between distributions into a single scalar, facilitating quantitative interpretation of the findings.
\end{itemize}

Therefore, we introduce a novel methodology to validate real and synthetic data similarity:
\begin{itemize}
    \item \textbf{Divergence-based Similarity Estimation:} Our approach leverages divergence estimation to quantify the discrepancy between the joint probability distributions of real and synthetic tabular data. This approach goes beyond traditional validation metrics, often focusing on marginal statistics.
    \item \textbf{Probabilistic Discriminator Network for Density Ratio Estimation:} To facilitate divergence calculations, we introduce an innovative approximation technique based on a probabilistic discriminator network \cite{tiao2018dre}. This network plays a crucial role in estimating the density ratio between real and synthetic distributions, which forms the basis for divergence computations.
    \item \textbf{KL Divergence and JS Divergence for Validation:} We acknowledge the widespread use of Kullback-Leibler (KL) divergence in information theory (often referred to as relative entropy) for divergence calculations. However, we recognize its limitations: KL divergence is not an actual distance in terms of mathematics \cite{cech1969} due to its asymmetry and the potential for infinite values in specific scenarios. To address this, we propose the inclusion of the Jensen-Shannon (JS) divergence, which offers the desirable properties of being bounded and symmetric, thus defining a proper distance metric. This bounded metric is crucial to assess the quality of synthetic data generation.
    \item \textbf{Rigorous Evaluation Framework:} We introduce a comprehensive evaluation framework to assess the efficacy of our proposed method. We initially apply our method to simple distributions where analytical solutions for divergences exist. This serves as a benchmark to validate the accuracy of our estimated divergences. Following the initial validation, we progress to more complex scenarios and finally apply our method to real-world datasets and their corresponding synthetically generated counterparts. This allows us to assess the effectiveness of our approach in practical applications.
    \item \textbf{Broad Applicability:} Due to the adaptability of our divergence-based validation method, it can be applied to various types of data beyond tabular formats. This has the potential to significantly impact numerous fields, such as healthcare, finance, and manufacturing. Adopting our approach can substantially improve the accuracy and reliability of synthetic data validation in various domains.
\end{itemize}

In summary, our contribution diverges from the traditional emphasis on divergence estimation, which is already a well-established and robust field of study. Instead, we focus on applying divergence estimation to validate synthetic data generation. As highlighted in our literature review, a notable opportunity exists to establish a standardized validation metric in this area. We propose that estimating divergences between real and synthetic data can serve as a definitive measure of the accuracy of the data generation process. By leveraging divergence metrics, we offer a quantitative measure that is both interpretable and grounded in solid theoretical principles. This approach introduces a robust metric that enhances the validation process for synthetic data generation, ensuring that synthetic data not only resemble the original in terms of individual feature distributions but also preserve the intricate interdependencies between features. These interdependencies are crucial for the data's authenticity and applicability in real-world scenarios. This advancement represents a significant step forward in standardizing validation practices and improving the reliability of synthetic datasets across various fields.

The structure of the paper is carefully designed to facilitate a comprehensive understanding of our proposal. It begins with Section I, an introductory section that sets the stage by highlighting the importance of robust validation metrics in the use of synthetic data across various domains. This is followed by Section II, a detailed background section that delves into the theoretical and analytical aspects of divergence measures, specifically the KL and JS divergences. Furthermore, this section discusses the density ratio estimation based on probabilistic classification, as outlined in reference \cite{tiao2018dre}, which is crucial for the subsequent divergence calculation. The methodology section (Section III) then defines our approach using generative models to obtain this density ratio estimation, detailing the process analytically, visually, and programmatically. The fourth section describes the experimental setup and discusses the validation results from four different experiments, each increasing in complexity, to demonstrate the efficacy of our approach. The paper concludes in Section V with a summary of our findings, highlighting key conclusions and potential future research lines. An appendix is also included, which presents additional results from our experiments, providing deeper insights into our validation process. 

\section{Background}
\label{sec:methods}
\subsection{Divergence definition}
Let us consider two probability distributions, $p(x)$ and $q(x)$, for a random variable $x$. In the context of probability distributions, divergence quantifies the dissimilarity between $p(x)$ and $q(x)$. In essence, it measures how different these distributions are regarding the probabilities they assign to each possible value of $x$. Divergences play a crucial role in various fields, including Machine Learning (ML), as they provide a way to compare and analyze the behavior of different probability distributions. 

The Kullback-Leibler divergence is a common measure in information theory to quantify the discrepancy between $p(x)$ and $q(x)$. Denoted $D_{\mathbb{KL}}\big(p(x)\|q(x)\big)$, the KL divergence captures the asymmetry of this difference by measuring the expected penalty incurred by assuming $q(x)$ to be an approximation of the true distribution $p(x)$. Due to its extensive applicability in various fields, the KL divergence remains a dominant metric for comparing probability distributions. The KL divergence between $p(x)$ and $q (x)$ is given by
\begin{equation}
    \label{eq:dkl}
    D_{\mathbb{KL}}\big(p(x) \| q(x)\big)=\int p(x) \log \frac{p(x)}{q(x)} d x.
\end{equation}

Unfortunately, this expression is usually intractable. Thus, we must approximate it using Monte Carlo (MC) simulation, assuming we can draw $L$ samples from $p(x)$:
\begin{equation}
\label{eq:dkl_montecarlo}
\int p(x) \log \frac{p(x)}{q(x)} d x \approx \frac{1}{L} \sum_{i=1}^L \log \frac{p\left(x_i\right)}{q\left(x_i\right)}, \quad x_i \sim p(x).
\end{equation}

To create a symmetric and bounded measure of divergence between $p(x)$ and $q(x)$, the Jensen-Shannon divergence is based on the KL divergence. The JS divergence denoted $D_{\mathbb{JS}}\big(p(x)\|q\big(x))\big)$, is a true distance defined as the KL average between both distributions and a common reference:
\begin{equation}
\label{eq:djs}
    \begin{aligned}
        &D_{\mathbb{JS}}\big(p(x) \| q(x)\big) \\ 
        &=\frac{1}{2} D_{\mathbb{KL}}\big(p(x)\|m(x)\big)
        +\frac{1}{2} D_{\mathbb{KL}}\big(q(x)\|m(x)\big).
    \end{aligned}
\end{equation}

The common reference distribution, $m(x)$, is typically chosen as the midpoint between the two original distributions, such as the average: $m(x)=\big(p(x)+q(x)\big)/2$. Notably, the JS divergence is inherently bounded. When employing a base-$2$ logarithm, the maximum divergence value is $1$.

Similarly to KL, the JS divergence can also be estimated using MC simulation:
\begin{equation}
    \label{eq:djs_montecarlo}
    \begin{aligned}
        D_{\mathbb{JS}}&\big(p(x)\|q(x)\big)\\
        \approx &\frac{1}{2L} \sum_{i=1}^L \log \frac{p\left(x_i\right)}{m\left(x_i\right)} +
        \frac{1}{2L} \sum_{i=1}^L \log \frac{q\left(\tilde{x}_i\right)}{m\left(\tilde{x}_i\right)}\\
        = &\frac{1}{2 L} \sum_{i=1}^L \log \left(\frac{2 p\left(x_i\right)}{p\left(x_i\right)+q\left(x_i\right)}\right)\\
        &+\frac{1}{2 L} \sum_{i=1}^L \log \left(\frac{2 q\left(\tilde{x}_i\right)}{p\left(\tilde{x}_i\right)+q\left(\tilde{x}_i\right)}\right),
    \end{aligned}
\end{equation}

assuming that we can generate $L$ samples from both distributions
\begin{equation}
    \left\{\begin{array}{l}
x_i \sim p(x) \\
\tilde{x}_i \sim q(x)
\end{array}\right ..
\end{equation}

\subsection{Estimating density ratio using probabilistic classification}
Density ratio estimation focuses on estimating the ratio $r^*(x)$ of two probability densities, $p(x)$ and $q(x)$, based solely on samples drawn from these distributions. This field has seen significant theoretical development, particularly regarding complex ratio estimators and their convergence properties \cite{Vapnik2000TheNO}. The ratio is therefore defined as
\begin{equation}
    r^*(x)=\frac{p(x)}{q(x)},
\end{equation}
where $*$ refers to the exact value. This approach avoids directly estimating individual densities since any errors in the denominator (i.e., $q(x)$) are dramatically amplified during the ratio calculation. The classical probabilistic classification approach remains popular among the various methods of estimating the density ratio due to its relative simplicity. This subsection reviews the method described by \cite{tiao2018dre} for estimating the KL divergence. We further demonstrate its application to the JS divergence estimation as well.

Estimating (\ref{eq:dkl}) and (\ref{eq:djs}) is based on determining the unknown density ratio $r^*$. This section establishes a connection between the density ratio of two distributions, $p(x)$ and $q(x)$, and a probabilistic classifier that optimally distinguishes samples drawn from these distributions. Consider a scenario in which we possess samples from both distributions, $p(x)$ and $q(x)$, with each sample labeled according to its origin. Using a classifier to estimate the class-membership probabilities for each sample, we can derive an estimator for the density ratio. Let $X_p = {x^{(1)}, x^{(2)}, ..., x^{(M_p)}}$ and $X_q = {\tilde{x}^{(1)}, \tilde{x}^{(2)}, ..., \tilde{x}^{(M_q)}}$ represent sets of real and synthetic data samples, respectively, where $M_p$ and $M_q$ denote the corresponding sample sizes. 

We proceed by forming a combined dataset, denoted as $D=(x_n, y_n)_{n=1}^N$, where $M_T=M_p+M_q$ represents the total number of samples. The label $y_n$ associated with each sample $x_n$ indicates its source distribution: $y_n=1$ for samples from $p(x)$ and $y_n=0$ for samples from $q(x)$. Consequently, we have $p(x)=\mathcal{P}(x \mid y=1)$ and $q(x)=\mathcal{P}(x \mid y=0)$, where $\mathcal{P}$ denotes the probability.

By applying Bayes' theorem, we can express the density ratio as:
\begin{equation}
    \begin{aligned}
        r^*(x) &= \frac{p(x)}{q(x)}=\frac{\mathcal{P}(x \mid y=1)}{\mathcal{P}(x \mid y=0)} \\
        &= \left(\frac{\mathcal{P}(y=1 \mid x) \mathcal{P}(x)}{\mathcal{P}(y=1)}\right)\left(\frac{\mathcal{P}(y=0 \mid x) \mathcal{P}(x)}{\mathcal{P}(y=0)}\right)^{-1} \\
        &= \frac{\mathcal{P}(y=0)}{\mathcal{P}(y=1)} \frac{\mathcal{P}(y=1 \mid x)}{\mathcal{P}(y=0 \mid x)}.
    \end{aligned}
\end{equation}

Equal prior probabilities are assumed for the source distributions, specifically $\mathcal{P}(y=0) = \mathcal{P}(y=1)$. This assumption is strategically made to prevent the estimation process from being skewed towards any particular class. By setting the prior probabilities to be equal, the ratio of marginal probabilities effectively cancels out, leaving:
\begin{equation}
    r^*(x) = \frac{\mathcal{P}(y = 1 \mid x)}{\mathcal{P}(y = 0 \mid x)}. 
    \label{eq:ratio}
\end{equation}

This expression reveals that $r^*$ can be estimated solely based on the posterior probability, $\mathcal{P}(y=1\mid x)$, which represents the probability that a sample originates from $p(x)$ given its features.

To simplify even more, the logit function, denoted by $\sigma^{-1}$, offers a convenient way to transform the posterior probability into the ratio of the original probabilities. The logit function is defined as the inverse of the logistic function $\sigma$:
\begin{equation}
    \sigma^{-1}(z) = \ln \left( \dfrac{z}{1-z} \right),
\end{equation}

where $z$ represents any probability. The logit function has the property that the log odds of two probabilities equal the log of their ratio. In our case, applying the logit function to both the numerator and denominator of the expression for (\ref{eq:ratio}) results in the log ratio of the original probabilities:

\begin{equation}
\begin{aligned}
&\ln \left( \dfrac{\mathcal{P}(y=1 \mid x)}{\mathcal{P}(y=0 \mid x)} \right) \\
&= \sigma^{-1} \left(\mathcal{P}(y = 1 \mid x)\right) - \sigma^{-1} \left(\mathcal{P}(y = 0 \mid x)\right)
\end{aligned}
\end{equation}

Since the prior probabilities are assumed to be equal, the second term cancels out, leaving us with:

\begin{equation}
\ln \left( r^*(x) \right) = \sigma^{-1} (\mathcal{P}(y = 1 \mid x))
\end{equation}

Taking the exponent of both sides recovers the original expression for $r^*(x)$ but expressed in terms of the odds of the posterior probability:

\begin{equation}
r^*(x) = \exp \left[ \sigma^{-1} (\mathcal{P}(y = 1 \mid x)) \right].
\end{equation}

Therefore, by applying the logit function, we can simplify the calculation of the density ratio. The transformation converts the posterior probability into a log-odds representation, ultimately leading to the desired log ratio. This establishes a crucial link between the density ratio and the probabilistic classifier. By training a classifier to effectively distinguish between samples from $p(x)$ and $q(x)$, we can obtain estimates of posterior probabilities and subsequently derive an estimate of the density ratio using the logit transformation.

\section{Methodology}
This research uses a neural network (NN) classifier, denoted $D_{\theta}$ with parameters $\theta$, to approximate the posterior probability $P(y=1\mid x)$.  This network acts as a discriminator to classify input samples originating from the $p(x)$ or $q(x)$ distributions. The network architecture is designed to capture the discriminative features that distinguish these distributions effectively. In this study, we aim to understand how input data variations influence the ratio estimation between distributions. To isolate the impact of input data, we deliberately chose not to fine-tune the model across different experiments, maintaining a consistent architecture throughout. This approach allows us to examine the robustness of the estimator across various data types without the confounding effects of model optimization. The model parameters were selected to balance interpretability and provide meaningful insights, ensuring that our results illustrate the estimator's practical capabilities rather than achieving optimal performance in each scenario. This strategy emphasizes the versatility and applicability of the estimator in real-world contexts, offering valuable insights into the challenges of synthetic data validation.%

Specifically, $D_{\theta}$ employs a three-layer architecture with a decreasing number of neurons per layer: 256 in the first hidden layer, followed by 64 and 32 neurons in the subsequent hidden layers, respectively. The Leaky Rectified Linear Unit activation function is used throughout the hidden layers to introduce nonlinearity into the model. In addition, dropout, batch normalization, and early stopping techniques are incorporated to prevent overfitting during the training process. 

The estimator for $r^*$ can be constructed as a function of the classifier's output:
\begin{equation}
    \begin{aligned}
        r_\theta(x) &=\exp \big[\sigma^{-1}\left(D_\theta(x)\right)\big] \\
        &\approx \exp \big[\sigma^{-1}(\mathcal{P}(y=1 \mid x))\big]=r^*(x).
    \end{aligned}
\end{equation}

The optimal class probability estimator is learned by minimizing a suitable loss function, such as the binary cross-entropy loss:
\begin{equation}
\begin{aligned} 
    &\mathcal{L}(\theta)
    \\&=-\mathbb{E}_{p(x)}\big[\log \mathcal{D}_\theta(x)\big]-\mathbb{E}_{q(x)}\big[\log (1-\mathcal{D}_\theta(x))\big].
\end{aligned}
\end{equation}

Once trained, the class probability estimator can construct MC estimates of the KL and JS divergences. The KL divergence between $p(x)$ and $q(x)$ can be estimated as
\begin{equation}
    \begin{aligned}
    D_{\mathbb{KL}}\big(p(x) \| q(x)\big)&=\mathbb{E}_{p(x)}\big[\log r^*(x)\big] \\
    &\approx \frac{1}{L} \sum_{i=1}^L \log r^*(x_i) 
    \approx \frac{1}{L} \sum_{i=1}^L \log r_\theta(x_i) \\
    &=\frac{1}{L} \sum_{i=1}^L \sigma^{-1}\big(\mathcal{D}_\theta\left(x_i\right)\big),
    \end{aligned}
\end{equation}

where $x_i \sim p(x)$, $L$ is the number of samples used for the estimate.

Similarly, the JS divergence can be estimated as:
\begin{equation}
    \begin{aligned}
        D_{\mathbb{JS}}&\big(p(x) \| q(x)\big) \\
        \approx& \frac{1}{2 L} \sum_{i=1}^L \log \big(2 \mathcal{D}_\theta(x_i)\big)\\
        &+\frac{1}{2 L} \sum_{i=1}^L \log \big(2-2 \mathcal{D}_\theta(\tilde{x}_i)\big) \\
        =&\frac{1}{2 L} \sum_{i=1}^L \log \big(\mathcal{D}_\theta(x_i)\big) \\
        &+\frac{1}{2 L} \sum_{i=1}^L \log \big(1-\mathcal{D}_\theta(\tilde{x}_i)\big),
    \end{aligned}
\end{equation}
where $\tilde{x}_i \sim q(x)$ denotes the number of samples used from the second distribution during divergence estimation.

Interestingly, the discriminator loss function, $\mathcal{L}(\theta)$, converges to a lower bound of the JS divergence up to a constant, $-2 \cdot D_{\mathbb{JS}}\big(p(x) \| q(x)\big)+\log 4$ \cite{tiao2018dre}. This relation can be established by analyzing the upper bound of the loss function:
\begin{equation}
\label{eq:upper_bound_loss}
    \begin{aligned}
        &\sup _\theta \mathbb{E}_{p(x)}\big[\log D_\theta(x)\big]+\mathbb{E}_{q(x)}\big[\log (1-D_\theta(x))\big] \\
        &=\mathbb{E}_{p(x)}[\log \mathcal{P}(y=1 \mid x)]+\mathbb{E}_{q(x)}[\log \mathcal{P}(y=0 \mid x)] \\
        &=\mathbb{E}_{p(x)}\big[\log \frac{p(x)}{p(x)+q(x)}\big]+\mathbb{E}_{q(x)}\big[\log \frac{q(x)}{p(x)+q(x)}\big] \\
        &=\mathbb{E}_{p(x)}\big[\log \frac{1}{2} \frac{p(x)}{m(x)}\big]+\mathbb{E}_{q(x)}\big[\log \frac{1}{2} \frac{q(x)}{m(x)}\big] \\
        &=\mathbb{E}_{p(x)}\big[\log \frac{p(x)}{m(x)}\big]+\mathbb{E}_{q(x)}\big[\log \frac{q(x)}{m(x)}\big]-2 \log 2 \\
        &=2 \cdot D_{\mathbb{JS}}\big(p(x) \| q(x)\big)-\log 4 .
    \end{aligned}
\end{equation}

Then, we can establish:
\begin{equation}
    \begin{aligned}
        &2 \cdot D_{\mathbb{JS}}\big(p(x) \| q(x)\big)-\log 4 \\
        &\geq \sup _\theta \mathbb{E}_{p(x)}\big[\log D_\theta(x)\big]+\mathbb{E}_{q(x)}\big[\log (1-D_\theta(x))\big].
    \end{aligned}
\end{equation}

Following this analysis, we can arrive at the desired inequality. However, we can manipulate the inequality by negating both sides for a more convenient form. 
\begin{equation}
\begin{aligned}
&-2 \cdot D_{\mathbb{JS}}\big(p(x) \| q(x)\big)+\log 4  \\
&\leq-\sup _\theta \mathbb{E}_{p(x)}\big[\log D_\theta(x)\big]+\mathbb{E}_{q(x)}\big[\log (1-D_\theta(x))\big] \\
 &=\inf _\theta-\mathbb{E}_{p(x)}\big[\log D_\theta(x)\big]-\mathbb{E}_{q(x)}\big[\log (1-D_\theta(x))\big] \\
 &=\inf _\theta \mathcal{L}(\theta).
\end{aligned}
\end{equation}

\begin{figure}[t]
    \includegraphics[width=0.5\textwidth]{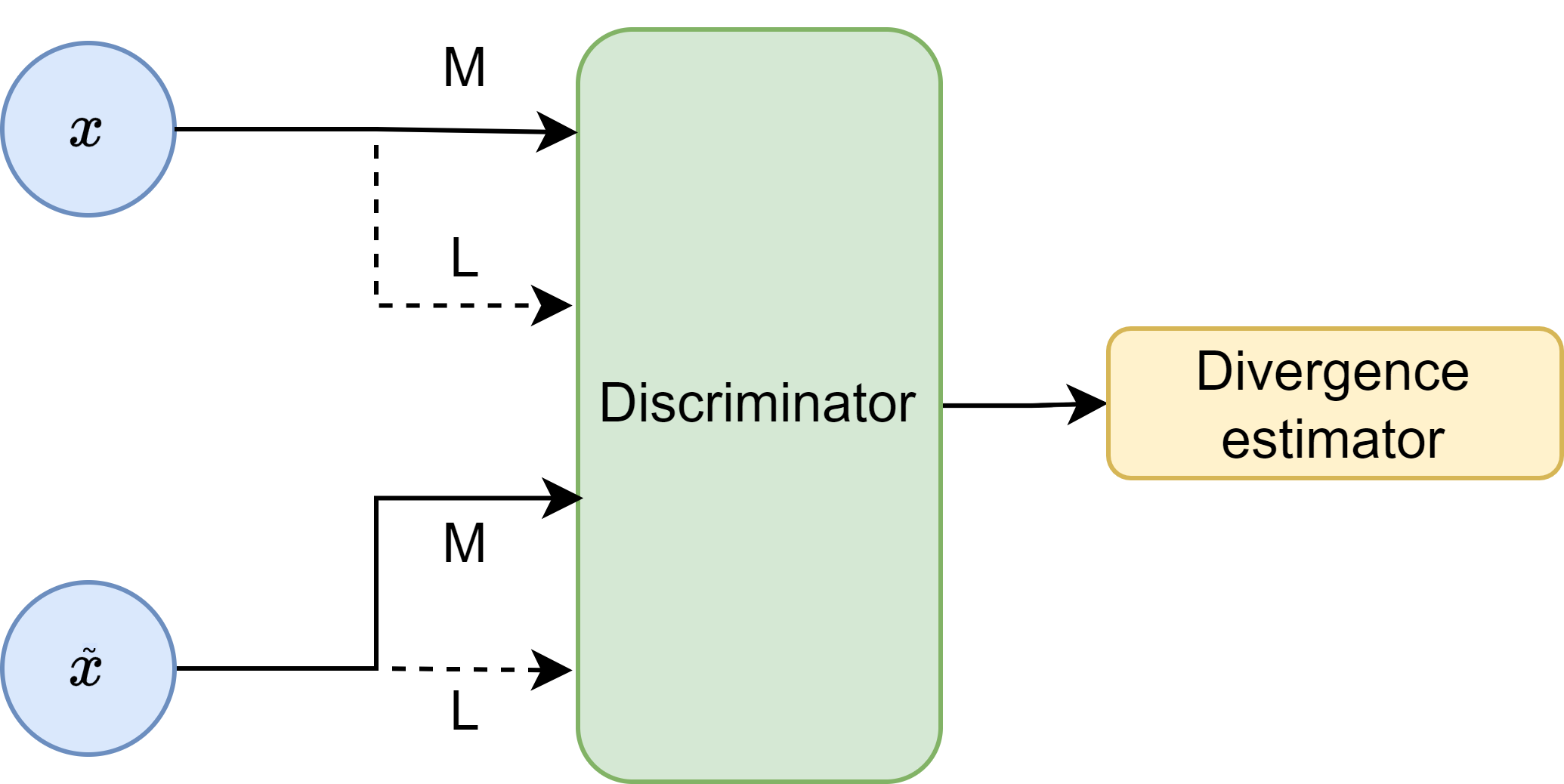}
    \centering
    \caption{Architecture of the neural network-based divergence estimator to assess the dissimilarity between samples from two datasets. The discriminator takes two sets of samples as input: $M$ samples from each set to train and $L$ samples from each to infer the divergence.}
    \label{fig:div_est_arch}
\end{figure}

The architecture of the divergence estimator proposed between two distributions is depicted in Fig. \ref{fig:div_est_arch}. The discriminator network receives two sets of samples: $M$ samples from the first distribution $p(x)$ labeled class $1$ and $M$ samples from the second distribution $q(x)$ labeled class $0$. During training, the discriminator aims to distinguish between these two sets. Subsequently, $L$ samples from each distribution are used to estimate the divergence between the underlying probability distributions.

\begin{figure}[b!]
    \includegraphics[width=\textwidth]{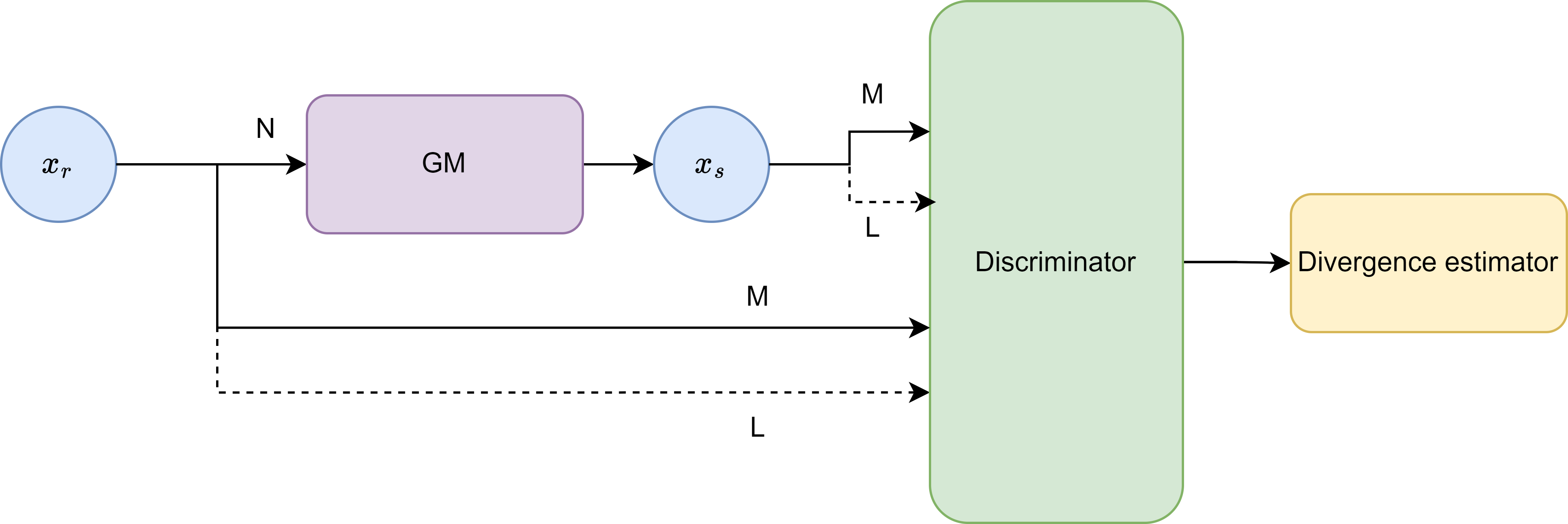}
    \centering
    \caption{Generative model for divergence estimation between real and synthetic data. The generative model learns an approximation of $N$ samples from real data $x_r$, denoted $x_s$. Subsequently, $M$ samples are drawn from each distribution to train the divergence estimation discriminator. Finally, $L$ samples from each distribution estimate the divergence between real and synthetic data.}
    \label{fig:gen_div}
\end{figure}

This study evaluates the dissimilarity between real and synthetic data generated by a Generative Model (GM). Minimizing the divergence between the real and synthetic data distributions is crucial in this context. Ideally, the divergence should approach zero, indicating that the generated data become virtually indistinguishable from the real data. This interchangeability allows the use of synthetic data in various applications where real data might be scarce or sensitive. From the discriminator's perspective, achieving minimal divergence implies that it cannot reliably differentiate between real and synthetic samples, indicating successful data generation. For GMs, the distributions of interest become:
\begin{itemize}
    \item $p_R(x)=p(x)$, representing the real data distribution.
    \item $p_S(x)=q(x)$, representing the synthetic data distribution generated by the model.
\end{itemize}

Training data consist of $N$ samples drawn from the real data distribution, denoted $x_r \sim p_R(x)$. These data are used to train the GM, as illustrated in Fig. \ref{fig:gen_div}. The GM learns to approximate the real data distribution as $p_S(x)$, allowing us to sample from this distribution and obtain synthetic data points $x_s \sim p_S(x)$. Following the architecture described in Fig. \ref{fig:div_est_arch}, these synthetic samples are then used along with the real data samples to train the discriminator and estimate the divergence between the real and synthetic data distributions.

\section{Experiments}
We aim to validate the effectiveness of our proposed divergence estimator for data generation. We achieve this through an exhaustive experimental evaluation in four different experiments, described in Table \ref{tab:experiments} designed to assess the estimator's performance under various conditions. The initial set of experiments focuses on controlled scenarios that involve comparisons of simple theoretical data distributions. Subsequently, the complexity is gradually increased by incorporating generative scenarios that reproduce real-world applications. This progression in complexity not only tests the estimator's ability to handle increasingly challenging conditions but also demonstrates its scalability and reliability. The increasing complexity is essential to ensure that our estimator remains accurate and efficient, even as the distributional assumptions become less defined and more reflective of the irregularities typically found in real data.%
\begin{table}[h!]
\caption{Experiments Configuration Summary. Experiments carried out to assess the effectiveness of the proposed approach. Experiments 1 and 2 investigate the impact of distribution complexity by varying the number of $M$ and $L$. Experiments 3 and 4 focus on the application to generative processes. A generative model is introduced to create synthetic data based on $N$ samples, enabling a comparison between real and synthetic scenarios.}
\renewcommand{\arraystretch}{1.25}
\centering
\resizebox{\textwidth}{!}{%
\begin{tabular}{c|cccccc}
\hline
\hline
\textit{\textbf{Experiment}} & \textbf{Compared Distributions}  & \textbf{Generative Model} & \textbf{N} & \textbf{M}  & \textbf{L}    \\
\hline
\hline
\textit{Experiment 1} & Multivariate Gaussian Distributions & -                      & -                           & [20, 200, 2000] & [20, 200, 2000] \\
\textit{Experiment 2} & Gaussian Mixture Distributions      & -                      & -                           & [20, 200, 2000] & [20, 200, 2000] \\
\textit{Experiment 3} & Gaussian Mixture and Synthetic Data & Gaussian Mixture Model & [10, 20, 30, ... , 150] & 2,000               & 2,000               \\
\textit{Experiment 4} & Real and Synthetic Data             & CTGAN\cite{xu2019modeling}, VAE\cite{apellaniz2024improved}             & 10,000                      & 7,500               & 1,000 \\
\hline
\end{tabular}}
\label{tab:experiments}
\end{table}

We comprehensively analyze the influence of sample size on the estimator's performance. This analysis covers the impact of the training and validation sets sizes for the discriminator network and the number of samples used during the generative process. To gain a deeper understanding, we evaluated different configurations for each parameter, as the table shows. For every experiment, all possible combinations of $N$, $M$, and $L$ are executed five times with unique random seeds. This rigorous approach employing multiple random seeds mitigates the effects of inherent randomness within training processes. Consequently, it provides a statistically robust evaluation and a more reliable representation of the observed trends. Ultimately, this comprehensive analysis facilitates identifying the configuration that yields the most accurate assessment of the divergence between data distributions. %

The inherent variability in data nature and complexity across different experiments needs diverse divergence estimation methods. Table \ref{tab:validation} details the specific techniques employed for each experiment, allowing for a comparative analysis whenever possible. The experiment's level of complexity guides the selection of the validation technique.
\begin{itemize}
    \item \textbf{Analytical Divergence:} This method, calculated only for $D_\mathbb{KL}$ for simplicity, represents the true divergence value due to the specific distributions used in the experiment.
    \item \textbf{MC Estimated Divergence:} This is a widely used estimation approach, but it can be computationally expensive.
    \item \textbf{Discriminator Estimated Divergence:} This method uses our proposed discriminator network to learn the density ratio between the two distributions and estimate the divergence.
\end{itemize}
\begin{itemize}
    \item \textbf{Analytical Divergence:} This method, calculated only for $D_\mathbb{KL}$ for simplicity, represents the true divergence value due to the specific distributions used in the experiment.
    \item \textbf{MC Estimated Divergence:} This is a widely used estimation approach, but it can be computationally expensive.
    \item \textbf{Discriminator Estimated Divergence:} This method uses our proposed discriminator network to learn the density ratio between the two distributions and estimate the divergence.
\end{itemize}
\begin{table}[h!]
\caption{Experiments Configuration Summary. Validation procedure for each experiment depending on their level of complexity.}
\renewcommand{\arraystretch}{1.25}
\centering
\resizebox{0.5\textwidth}{!}{%
\begin{tabular}{c|c}
\hline
\hline
\textit{\textbf{Experiment}} & \textbf{Validation Technique}\\
\hline
\hline
\textit{Experiment 1} & Analytical, MC and Discriminator Estimations \\
\textit{Experiment 2} & MC and Discriminator Estimations \\
\textit{Experiment 3} & MC and Discriminator Estimations \\
\textit{Experiment 4} & Discriminator Estimation \\
\hline
\end{tabular}}
\label{tab:validation}
\end{table}

In the experiments conducted, the parameter values of the different models used were explicitly chosen to yield illustrative results to analyze the performance of the proposed estimator in validating synthetic data generation. We recognize that alternative parameters could be employed or fine-tuning could be undertaken to identify optimal settings; however, we intend to observe the effect of the input data on the models as such and to demonstrate their potential. This approach allows us to focus on understanding how variations in input data influence the estimator's ability to validate synthetic datasets, thereby showcasing the practical applicability and robustness of the estimator across different data scenarios.%

For complete transparency and reproducibility, the data and code used in this study are publicly available on our repository (\url{https://github.com/Patricia-A-Apellaniz/divergence_estimator}). 

\subsection{Experiment 1}
\begin{figure}[h!]
    \includegraphics[width=\textwidth]{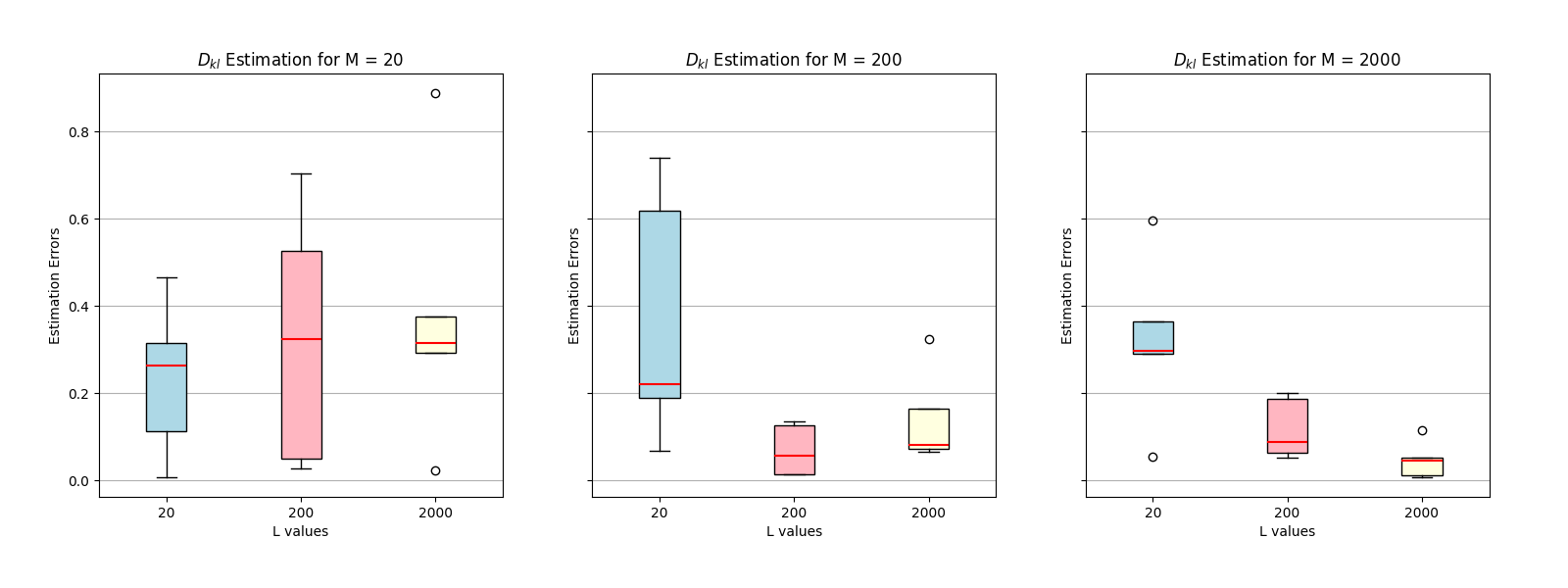}
    \centering
    \caption{Estimation error representation for $D_\mathbb{KL}$ in Experiment 1. Results are shown for different combinations of training sample sizes $M$ and validation sample sizes $L$. As expected, a decrease and precision in the error is observed with increasing values of $M$ and $L$.}
    \label{fig:error_kl_divergences_use_case_1}
\end{figure}

\begin{figure}[h!]
    \includegraphics[width=\textwidth]{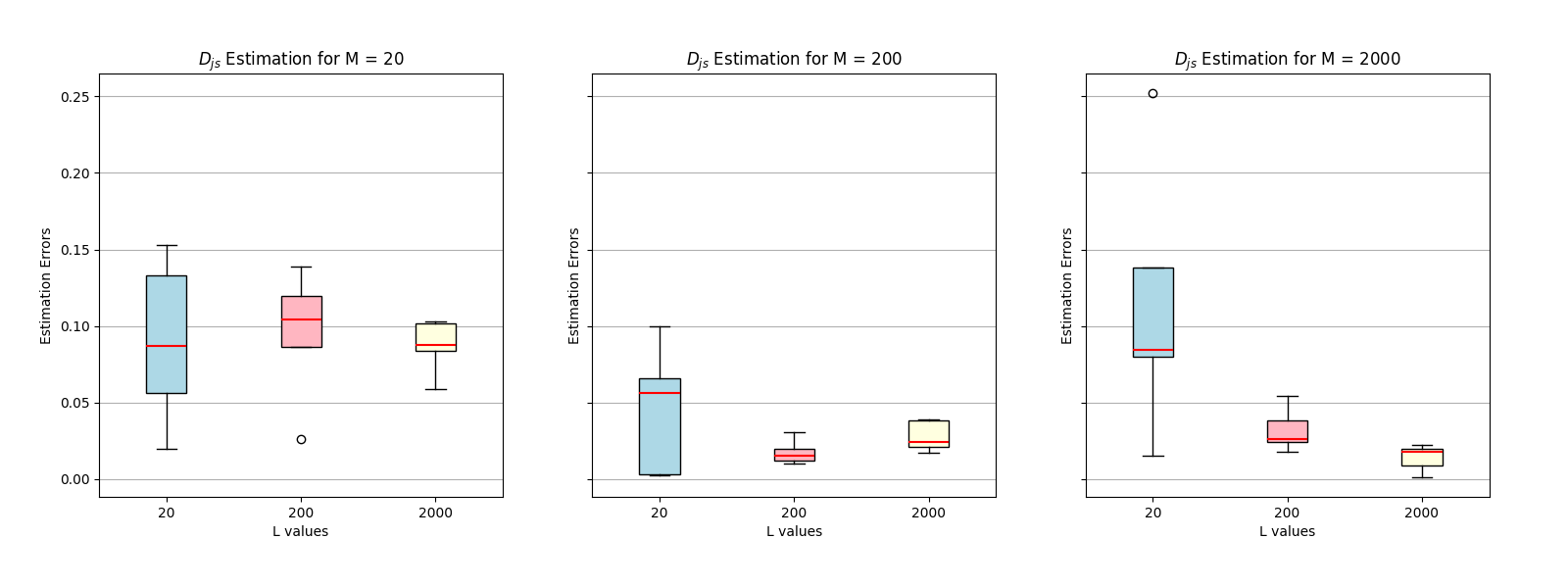}
    \centering
    \caption{Estimation error representation for $D_\mathbb{JS}$ in Experiment 1. Results are shown for different combinations of training sample sizes $M$ and validation sample sizes $L$. As expected, a decrease and precision in the error is observed with increasing values of $M$ and $L$.}
    \label{fig:error_js_divergences_use_case_1}
\end{figure}

\begin{figure}[h!]
    \includegraphics[width=\textwidth]{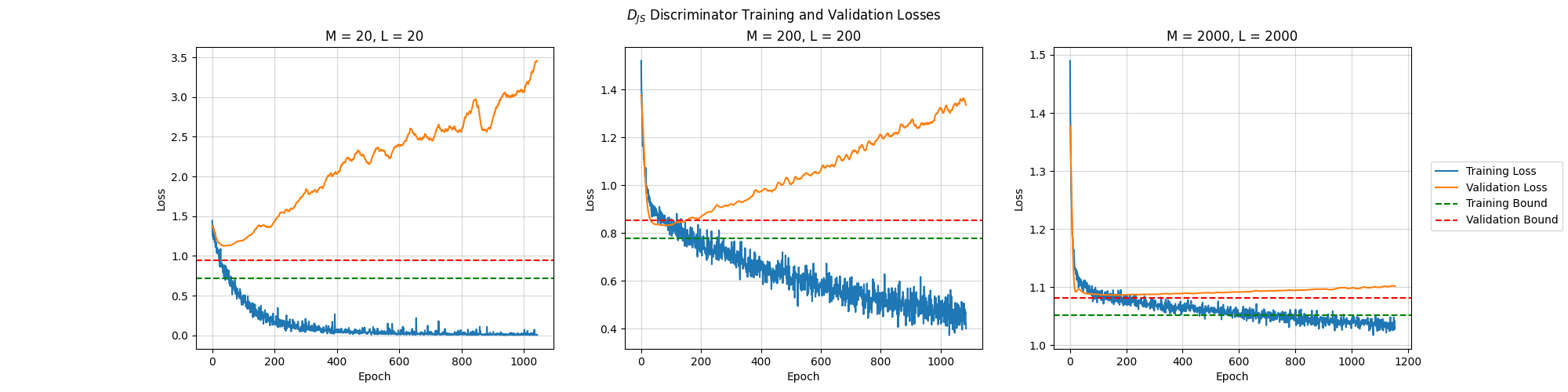}
    \centering
    \caption{Discriminator loss curves for Experiment 1. The loss curves show a clear overfitting due to low sample sizes. Green and red dashed lines represent theoretical convergence values.}
    \label{fig:JS_discriminator_losses}
\end{figure}

Table \ref{tab:use_case_1_div_values} presents the results obtained for $D_{\mathbb{KL}}$ and $D_{\mathbb{JS}}$ between random multivariate Gaussian distributions. A significant advantage of this study is the availability of a closed-form solution for the KL divergence, which serves as the ground truth value. We use the MC simulation estimate for the JS divergence as the ground truth. To ensure the reliability of this experiment, we opted for multivariate Gaussian distributions with a dimensionality of 10. This allows us to compute the true divergence values and assess the accuracy of estimating the ratio of the proposed discriminator. The results reveal a critical relation between sample size and estimation accuracy, particularly for the discriminator-based approach. When the number of samples is limited ($M = 20$, $L = 20, 200$), the estimated divergence from the discriminator deviates significantly from both the analytical value and the MC estimate. This suggests a potential overfitting due to insufficient data, leading to underestimating the true divergence. However, as the number of samples increases, the estimated divergence from the discriminator network progressively converges towards the ground truth value, followed by improved confidence interval precision. Figs. \ref{fig:error_kl_divergences_use_case_1} and \ref{fig:error_js_divergences_use_case_1} confirm this trend. Both figures illustrate how increasing the number of training samples $M$ and validation samples $L$ reduces the error associated with the estimated divergence ratio. Further support for the correlation between sample size and estimation accuracy comes from the discriminator loss curves depicted in Fig. \ref{fig:JS_discriminator_losses}. As detailed in \cite{tiao2018dre}, the loss function converges to a constant value approximating the Jensen-Shannon divergence. The figure visually confirms this concept, as the loss curves flatten with increasing training $M$ and validation $L$ samples. In contrast, a low number of samples (e.g., $M=20$, $L=20$) results in larger fluctuations in the loss function, potentially indicating discriminator overfitting. These validations demonstrate the effectiveness of our proposed method: with sufficient training data, the discriminator can accurately learn the density ratio and provide reliable estimates of the divergences, particularly for the JS divergence.

\begin{table*}[h!]
\caption{Impact of Training and Validating Samples on $D_\mathbb{KL}$ and  $D_\mathbb{JS}$ Estimation for Experiment 1. Analytical $D_\mathbb{KL}$ along with MC $D_\mathbb{KL}$ and $D_\mathbb{JS}$ estimations, as well as proposed discriminator estimations for both divergences. Results are displayed for various combinations of training samples $M$ and validation samples $L$. There is a clear correlation between the number of samples used and the estimation error.}
\renewcommand{\arraystretch}{1.25} 
\centering
\resizebox{\textwidth}{!}{%
\begin{tabular}{cc|ccc|cc}
\hline
\hline
\textbf{M} 
& \textbf{L} 
& \textbf{Analytical $D_\mathbb{KL}$} 
&  \textbf{MC Estimated $D_\mathbb{KL}$}
& \textbf{Discriminator Estimated $D_\mathbb{KL}$}
& \textbf{MC Estimated $D_\mathbb{JS}$}
& \textbf{Discriminator Estimated $D_\mathbb{JS}$} \\
\hline
\hline
\multirow{3}{*}{20}  & 20  & \multirow{9}{*}{1.035}    & $1.168\pm 0.203$     & $0.802\pm 0.152$   & $0.320\pm 0.105$     & $0.213\pm 0.072$\\
                    & 200   &        & $1.026\pm 0.221$      & $0.709\pm 0.248$     & $0.301\pm 0.055$     & $0.200\pm 0.038$\\
                      & 2000   &    & $1.024\pm 0.193$   & $0.665\pm 0.084$ & $0.288\pm 0.047$     & $0.209\pm 0.028$ \\
\cline{1-2}
\cdashline{3-3}
\cline{4-7}
\multirow{3}{*}{200}  & 20     &         & $1.221\pm 0.422$    & $0.991\pm 0.520$   & $0.320\pm 0.055$     & $0.289\pm 0.053$ \\
                      & 200     &      & $1.114\pm 0.257$     & $1.099\pm 0.227$  & $0.313\pm 0.055$     & $0.294\pm 0.049$  \\
                      & 2000      &         & $1.026\pm 0.191$   & $0.955\pm 0.101$  & $0.289\pm 0.038$     & $0.267\pm 0.038$  \\
\cline{1-2}
\cdashline{3-3}
\cline{4-7}
\multirow{3}{*}{2000} & 20      &         & $0.841\pm 0.190$    & $0.833\pm 0.229$  & $0.219\pm 0.113$     & $0.218\pm 0.111$    \\
                      & 200    &           & $1.004\pm 0.190$    & $0.978\pm 0.185$  & $0.300\pm 0.047$     & $0.290\pm 0.049$   \\
                      & 2000       &        & $1.055\pm 0.212$   & $1.060\pm 0.200$      & $0.299\pm 0.051$     & $0.293\pm 0.050$   \\                             
\hline
\end{tabular}}
\label{tab:use_case_1_div_values}
\end{table*}

\begin{figure}[t!]
    \includegraphics[width=\textwidth]{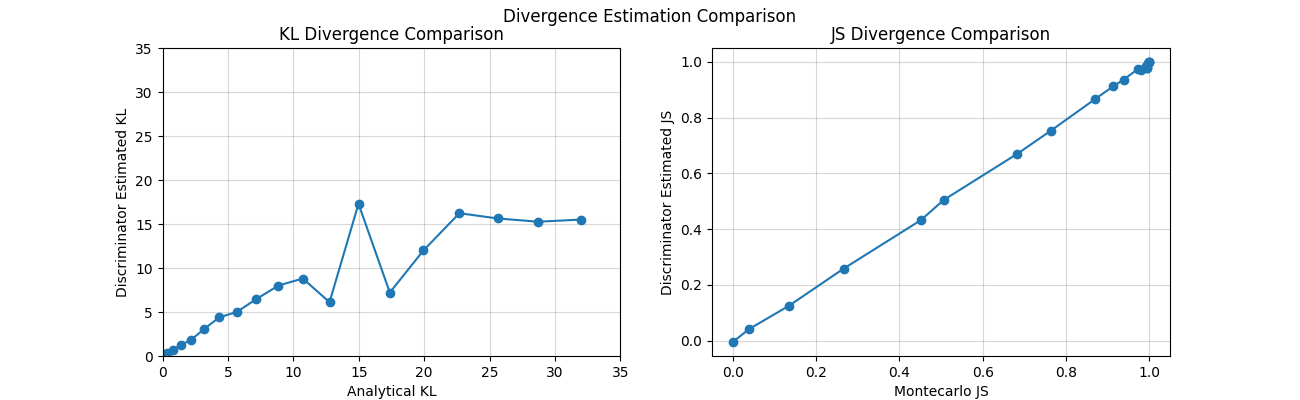}
    \centering
    \caption{Relation between ground truth and estimated divergences as distribution separation increases in Experiment 1.}
    \label{fig:deviations}
\end{figure}

Leveraging Experiment 1, we further emphasize the superior robustness of our proposed approach in estimating the JS divergence, regardless of variations in distribution separation. Fig. \ref{fig:deviations} illustrates the relation between estimated divergences and ground truth values as the separation between two 4-dimensional multivariate Gaussian distributions increases. As observed, the relation for the JS divergence remains almost linear until it approaches 1. While the relation appears less linear for values near 1, it is essential to note that the primary application of this discriminator lies in comparing similar distributions. Therefore, the focus should be on the performance of low JS divergences. Meanwhile, the KL divergence exhibits a nonlinear relation starting from the first comparisons, with the estimated KL deviating from the ground truth probably because of bias introduced by the discriminator. 

\subsection{Experiment 2}
This experiment investigates the performance of the proposed divergence estimation method in Gaussian mixture distributions. These distributions offer a higher complexity level than those used in Experiment 1. Each Gaussian mixture distribution comprises two independent isotropic Gaussian components with distinct mixing probabilities (weights assigned to each component within the mixture). Unlike the prior case, where analytical ground truth for divergence was obtainable, we rely solely on MC approximations as ground truth here and in the following experiment. Similar results were obtained for this experiment compared to the first. 

Tables \ref{tab:dkl_values_second_use_case} and \ref{tab:djs_values_second_use_case} focus on a practical range for training sample sizes $M$ and validation sample sizes $L$ (both set to $200, 2000$) and present the estimated divergences obtained using our proposed method compared to MC estimates. The results demonstrate that, for both divergences, the proposed method achieves results comparable to the MC estimates. These findings further support the previously observed correlation between sample size and estimation accuracy. Additionally, we observe a narrowing of confidence intervals with increasing sample sizes for both divergence measures. This trend indicates that the estimation becomes progressively more precise as the amount of available data increases. Overall, this experiment's results underscore the proposed method's effectiveness in estimating divergences for more intricate distributions, such as Gaussian mixtures. The remaining analysis can be found in the Appendix.

\begin{table}[h!]
\caption{Impact of Training and Validating Samples on $D_\mathbb{KL}$ Estimation for Experiment 2. MC and proposed method $D_\mathbb{KL}$ estimation for various combinations of training and validation samples $M$ and $L$. The results highlight a clear correlation between the number of samples used and the estimation error.}
\renewcommand{\arraystretch}{1.25} 
\centering
\resizebox{0.5\textwidth}{!}{%
\begin{tabular}{cc|cc}
\hline
\hline
\textbf{M} 
& \textbf{L} 
& \multicolumn{1}{c}{\textbf{MC Estimated $D_\mathbb{KL}$}} 
& \multicolumn{1}{c}{\textbf{Discriminator Estimated $D_\mathbb{KL}$}} \\
\hline
\hline
\multirow{2}{*}{200}   & 200    & $2.669\pm 0.118$     & $2.686\pm 0.294$     \\
                      & 2000   & $2.887\pm 0.045$     & $3.037\pm 0.517$  \\
\hline
\multirow{2}{*}{2000}  & 200    & $2.710\pm 0.100$     & $2.628\pm 0.357$    \\
                      & 2000   & $2.850\pm 0.071$     & $2.913\pm 0.129$    \\                            
\hline
\end{tabular}}
\label{tab:dkl_values_second_use_case}
\end{table}

\begin{table}[h!]
\caption{Impact of Training and Validating Samples on $D_\mathbb{JS}$ Estimation for Experiment 2. MC and proposed method $D_\mathbb{JS}$ estimation for various combinations of training and validation samples $M$ and $L$. The results highlight a clear correlation between the number of samples used and the estimation error.}
\renewcommand{\arraystretch}{1.25} 
\centering
\resizebox{0.5\textwidth}{!}{%
\begin{tabular}{cc|cc}
\hline
\hline
\textbf{M} 
& \textbf{L} 
& \multicolumn{1}{c}{\textbf{MC Estimated $D_\mathbb{JS}$}} 
& \multicolumn{1}{c}{\textbf{Discriminator Estimated $D_\mathbb{JS}$}} \\
\hline
\hline
\multirow{2}{*}{200}   & 200    & $0.392\pm 0.013$     & $0.374\pm 0.020$     \\
                      & 2000   & $0.428\pm 0.009$     & $0.412\pm 0.009$  \\
\hline
\multirow{2}{*}{2000}  & 200    & $0.415\pm 0.020$     & $0.412\pm 0.019$    \\
                      & 2000   & $0.423\pm 0.014$     & $0.420\pm 0.013$    \\                              
\hline
\end{tabular}}
\label{tab:djs_values_second_use_case}
\end{table}

\subsection{Experiment 3}
\begin{figure}[t!]
    \includegraphics[width=\textwidth]{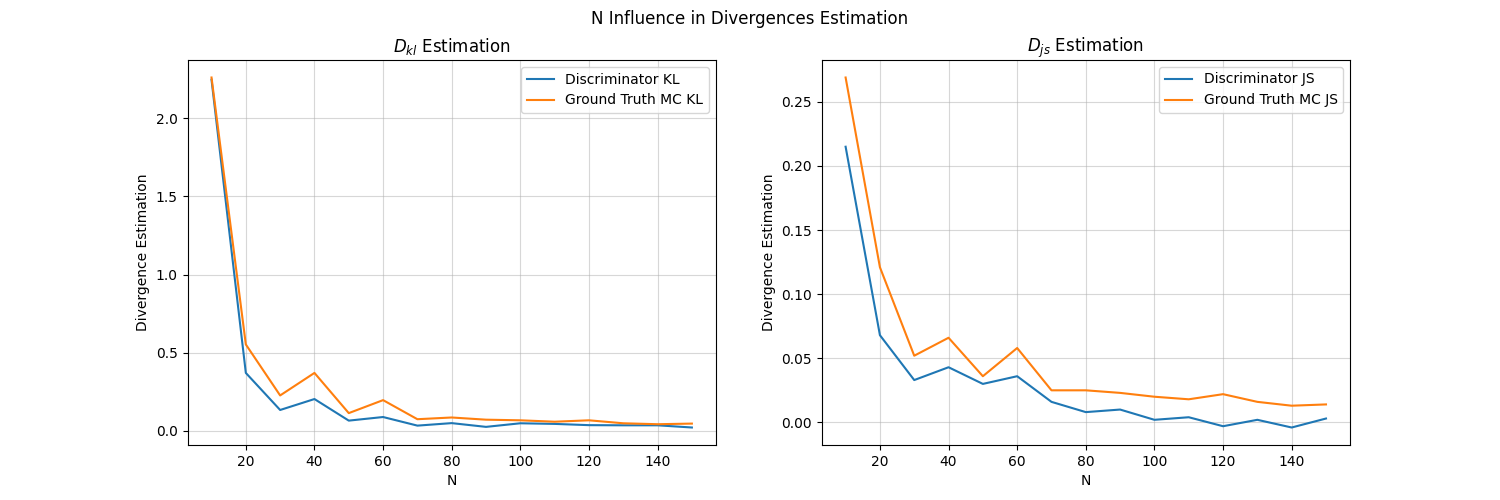}
    \centering
    \caption{Estimated $D_\mathbb{KL}$ and $D_\mathbb{JS}$ compared to ground truth values for varying sample sizes of $N$ in Experiment 3. Note the independent y-axes for each divergence measure due to their inherent scale differences, with $D_\mathbb{JS}$ exhibiting lower values than $D_\mathbb{KL}$.}
    \label{fig:use_case_3_n}
\end{figure}
The following experiments introduce the concept of a generative process, where synthetic data are generated based on $N$ samples from a particular data distribution. We investigate the influence of the quality of the GM in terms of the number of samples used to generate. This specific experiment uses a Gaussian mixture with two components as the real distribution. We employ a Gaussian mixture model (GMM) as the GM. The trained GMM then generates synthetic data that serve as an approximation of the real data. Finally, we estimate the divergence using our approach with a sufficient number of samples to achieve a low confidence interval and compare it to the MC simulation estimation. We analyze the behavior of the GMM under different training configurations, varying the number of training samples $N$. Since previous experiments demonstrated that a high number of samples for both $M$ and $L$ achieve better divergence estimations, we have fixed $M=2000$ and $L=2000$ for this experiment. We compare the estimated divergence errors for this combination of training samples $M$ and validation $L$ used by the divergence estimator. 

Fig. \ref{fig:use_case_3_n} shows the estimated divergences as $N$ increases. The results demonstrate a clear correlation between the size of the training data for the GMM, $N$, and the estimated divergence errors. When the GMM is poorly trained due to a limited number of samples ($N=10$), it cannot effectively capture the underlying data patterns. This leads to generated data that significantly deviate from the real distribution. Consequently, the real divergence values increase significantly, and a large estimation error occurs, particularly for unbounded $D_\mathbb{KL}$. $D_\mathbb{JS}$, which is inherently bounded between 0 and 1, exhibits less extreme error values. As the number of training samples for the GMM increases ($N=200$ and $N=2000$), the generated data become more representative of the real distribution. This results in lower divergence values and reduced estimation errors for $D_\mathbb{KL}$ and $D_\mathbb{JS}$. This finding supports the validity of our proposed validation technique for generative processes involving Gaussian mixture distributions.

\subsection{Experiment 4}
The final experiment introduces a more realistic setting by incorporating a real-world dataset along with its synthetic counterpart generated using a GM. This scenario retains the concept of a generative process, but the target distribution becomes the real-world data itself. The dataset chosen is Adult. As described in \cite{adult}, it contains information on $32,561$ individuals used to predict their annual income exceeding \$50,000. It represents an extract from the $1994$ U.S. Census database \cite{census}. For our experiments and after the previous analysis was done, we used a subset of $10,000$ samples obtained from \cite{sdv}. This dataset comprises 14 features encompassing various data types, including categorical, binary, and integer values. Based on the previous analysis of the influence of the sample size, we fixed the hyperparameters for this experiment at $N=10000$, $M=7500$, and $L=1000$. To evaluate GMs, we compared the performance of two state-of-the-art approaches: the widely used CTGAN\cite{xu2019modeling} and the VAE-based generator proposed in \cite{apellaniz2024improved} (VAE), which achieved superior results based on several validation criteria. 

Table \ref{tab:real_scenario_results} confirms this, obtaining lower divergence results for \cite{apellaniz2024improved}. To explain why the VAE may outperform the CTGAN in our experiments, we refer to \cite{apellaniz2024improved}, which compares the VAE generator with CTGAN and TVAE using metrics similar to ours. The results support our observations. We suggest that VAEs perform better due to their stability in managing complex data distributions, unlike GANs, which struggle with convergence. This stability in VAEs contributes to their enhanced ability to generate more realistic synthetic data, highlighting significant differences in model architecture and distribution handling. This experiment focuses on evaluating the ability of the divergence estimator to assess the similarity between the real data and its synthetic counterpart. This experiment provides valuable insights into the practical application of the proposed method in real-world scenarios where measuring the similarity between real and synthetic data is crucial. 
\begin{table}[h!]
\caption{Comparison of Generative Models for Real-World Data. Similarity comparison between real data and their synthetic counterparts generated by each model using estimated divergences, $D_\mathbb{KL}$ and $D_\mathbb{JS}$. Lower values in both metrics indicate a greater similarity between real and synthetic data.}
\renewcommand{\arraystretch}{1.5} 
\centering
\resizebox{0.5\textwidth}{!}{%
\begin{tabular}{cc|cc}
\hline
\hline            
\textbf{CTGAN $D_\mathbb{KL}$ }
& \textbf{VAE $D_\mathbb{KL}$} 
& \textbf{CTGAN $D_\mathbb{JS}$} 
& \textbf{VAE $D_\mathbb{JS}$} \\ 
\hline
 $0.342 \pm 0.016$     
 & $0.185\pm 0.030$    
 & $0.138 \pm 0.002$     
 & $0.099 \pm 0.002$   \\
\hline
\end{tabular}}
\label{tab:real_scenario_results}
\end{table}

\section{Conclusion}
This research proposes a novel and practical approach for validating synthetic tabular data generated by various models. The core of this method lies in using a divergence estimator based on a probabilistic classifier to capture the discrepancies between the real and synthetic data distributions. This approach overcomes the limitations associated with traditional marginal divergence comparisons by considering the joint distribution. While marginal comparisons assess the similarity of individual features between real and synthetic data, they can be misleading. Even if the marginal distributions of each feature appear similar, the joint distribution, which captures the relations between features, may differ significantly. This can lead to unrealistic synthetic data, where individual features appear plausible, but their co-occurrences do not represent the real data. By considering the joint distribution, our proposed method provides a more comprehensive assessment of the quality of synthetic tabular data.

The efficacy of the proposed method is comprehensively evaluated through a series of experiments with progressively increasing complexity. The initial phase establishes a solid foundation by analyzing the performance in controlled scenarios with well-defined theoretical distributions. The results demonstrate that the accuracy of divergence estimates is highly dependent on the amount of training data available for both the GM and the divergence estimator network. Subsequent experiments explore more intricate scenarios involving Gaussian mixture distributions and real-world datasets. The findings consistently support the effectiveness of the proposed method in approximating the true divergences. Our emphasis in this paper has been to show the advantages of using divergences as synthetic data validation for tabular data.

This research offers significant contributions that extend beyond the specific tabular data validation domain. The proposed methodology facilitates better validation practices for various fields dependent on GMs. Its key strength lies in capturing complex relations between different data distributions, leading to more robust and reliable validation processes. In addition to the positive results, the study also highlights the importance of the quality of the GM. When it is inadequately trained due to insufficient data, it can significantly impact the accuracy of the divergence estimation. This emphasizes the need to consider GM training procedures carefully to ensure reliable validation results.

Several promising avenues exist for future research. One direction involves exploring the potential to extend the proposed approach to more complex data structures, such as images or time series data. Additionally, investigating the integration of this validation technique within GM training pipelines could enable the development of self-improving GMs that can automatically adjust their parameters to generate data that closely resemble the target distribution. Moreover, addressing the impact of changing the assumption of prior probabilities for the source distributions to be compared, where the number of real and synthetic samples may vary, presents a significant challenge. Techniques to mitigate the effects of unbalanced classes in the regression framework could be explored. These techniques may include adaptive sampling strategies, class weighting, or methods specifically tailored to handle imbalanced data distributions. Implementing such approaches could enhance the robustness and applicability of the validation technique across diverse datasets and real-world applications. Finally, developing a robust methodology to estimate divergences when limited samples are available presents a crucial challenge. Addressing this challenge would further enhance the versatility and practicality of the proposed validation technique in real-world scenarios with restricted data availability. Furthermore, it would be beneficial to investigate alternative density ratio estimation techniques, such as those presented in \cite{pmlr-v151-choi22a, rhodes2020telescoping}. These methods may offer advantages regarding sample size requirements, estimation error, or other relevant properties. 

\section*{Acknowledgments}
This research was supported by GenoMed4All and SYNTHEMA projects. Both have received funding from the European Union’s Horizon 2020 research and innovation program under grant agreement No 101017549 and 101095530, respectively. The authors declare that they have no known competing financial interests or personal relations that could have appeared to influence the work reported in this paper.

\section*{Appendix}\label{sec:app}
Leveraging the validation process established in Experiment 1, Experiment 2 employs the same methodology. The remaining analyses for Experiment 2 are presented as follows.

Table \ref{tab:use_case_2_div_values} extends the findings of the previous experiment by investigating the performance of our proposed approach with a restricted number of training and validation samples for the divergence estimator. The results indicate a trend of convergence towards the known ground truth divergence values as the number of samples utilized for both training and validation is progressively increased.

\begin{figure}[h!]
    \includegraphics[width=\textwidth]{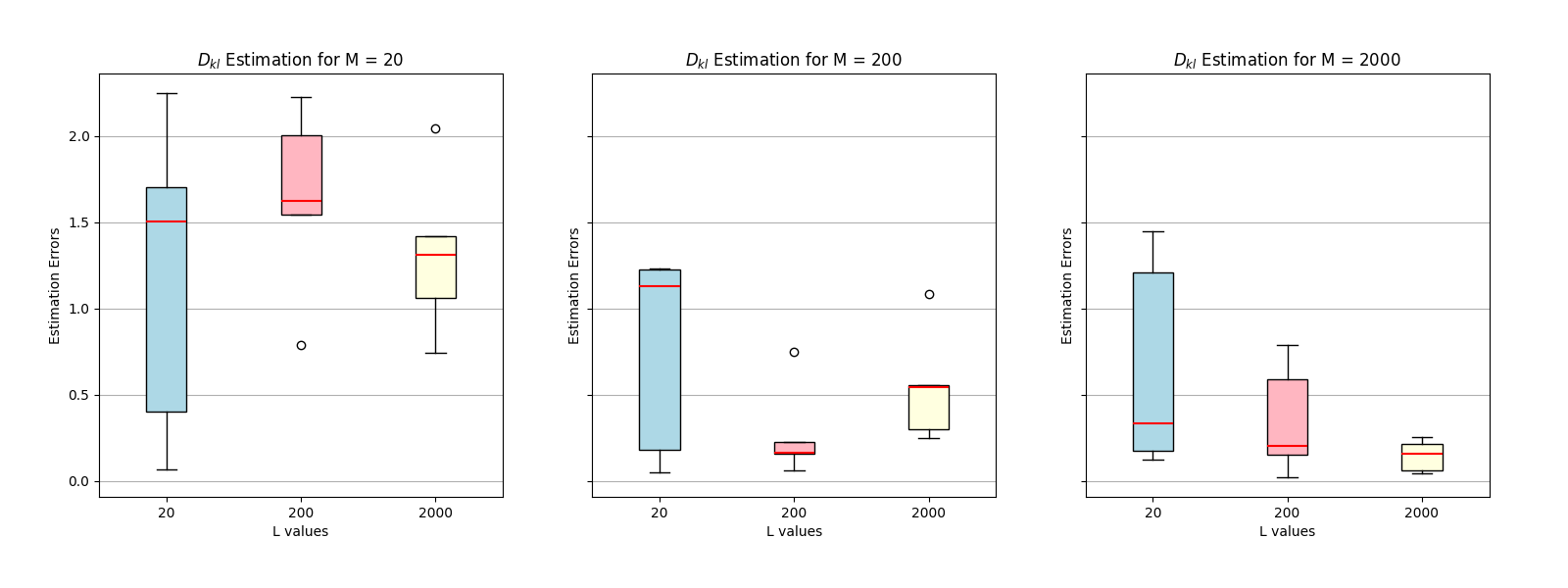}
    \centering
    \caption{Estimation error representation for $D_\mathbb{KL}$ in Experiment 2. Results are shown for different combinations of training sample sizes $M$ and validation sample sizes $L$. As expected, a decrease and precision in the error is observed with increasing values of $M$ and $L$.}
    \label{fig:error_kl_divergences_use_case_2}
\end{figure}

\begin{figure}[h!]
    \includegraphics[width=\textwidth]{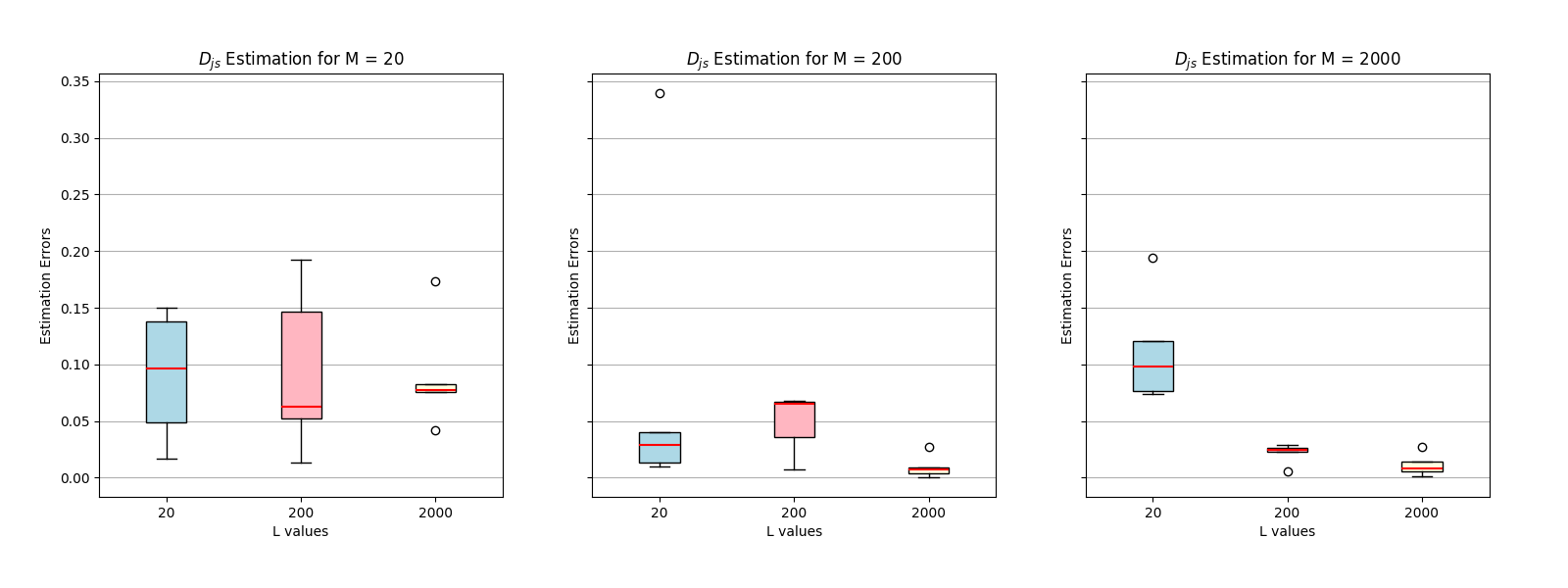}
    \centering
    \caption{Estimation error representation for $D_\mathbb{JS}$ in Experiment 2. Results are shown for different combinations of training sample sizes $M$ and validation sample sizes $L$. As expected, a decrease and precision in the error is observed with increasing values of $M$ and $L$.}
    \label{fig:error_js_divergences_use_case_2}
\end{figure}

Figs. \ref{fig:error_kl_divergences_use_case_2} and \ref{fig:error_js_divergences_use_case_2} provide further substantiation for this observed trend.

\begin{figure}[h!]
    \includegraphics[width=\textwidth]{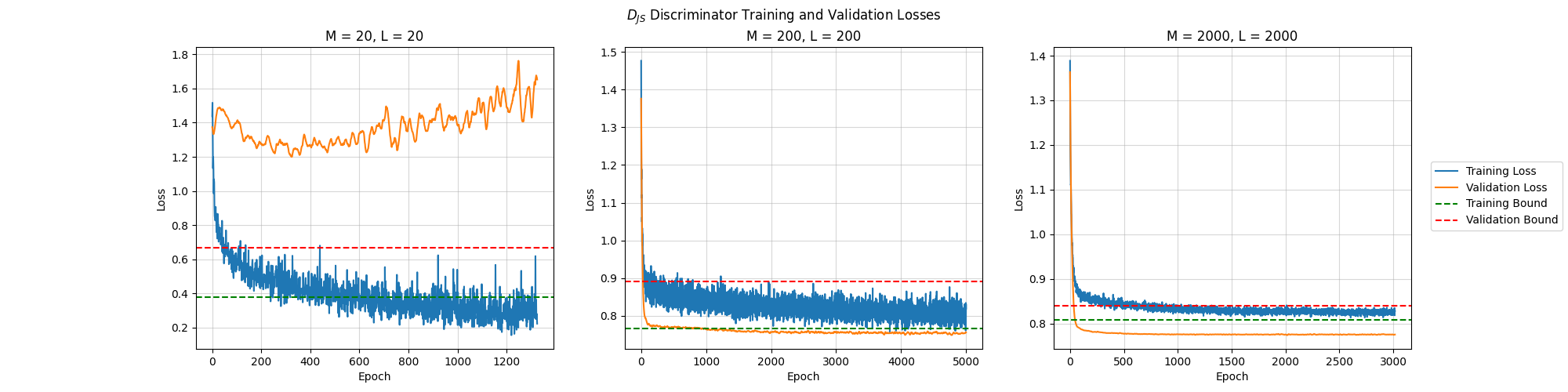}
    \centering
    \caption{Discriminator loss curves for Experiment 2. The loss curves show a clear overfitting due to low sample sizes. Green and red dashed lines represent theoretical convergence values.}
    \label{fig:JS_discriminator_losses_use_case_2}
\end{figure}

Additionally, Fig. \ref{fig:JS_discriminator_losses_use_case_2} delves into the behavior of the discriminator's loss function across various sample size combinations.

\begin{table}[h!]
\caption{Impact of Training and Validating Samples on $D_\mathbb{KL}$ and  $D_\mathbb{JS}$ Estimation for Experiment 2. Analytical $D_\mathbb{KL}$ along with MC $D_\mathbb{KL}$ and $D_\mathbb{JS}$ estimations, as well as proposed discriminator estimations for both divergences. Results are displayed for various combinations of training samples $M$ and validation samples $L$. There is a clear correlation between the number of samples used and the estimation error.}
\renewcommand{\arraystretch}{1.25} 
\centering
\resizebox{\textwidth}{!}{%
\begin{tabular}{cc|cc|cc}
\hline
\hline
\textbf{M} 
& \textbf{L} 
&  \textbf{MC Estimated $D_\mathbb{KL}$}
& \textbf{Discriminator Estimated $D_\mathbb{KL}$}
& \textbf{MC Estimated $D_\mathbb{JS}$}
& \textbf{Discriminator Estimated $D_\mathbb{JS}$} \\
\hline
\hline
\multirow{3}{*}{20}  & 20     & $3.245\pm 0.820$     & $1.832\pm 0.912$   & $0.490\pm 0.083$     & $0.371\pm 0.078$\\
                    & 200   &       $2.699\pm 0.095$      & $1.193\pm 0.430$     & $0.439\pm 0.017$     & $0.328\pm 0.058$\\
                      & 2000     & $2.833\pm 0.032$   & $1.512\pm 0.381$ & $0.423\pm 0.006$     & $0.331\pm 0.039$ \\
\cline{1-6}
\multirow{3}{*}{200}  & 20           & $2.639\pm 0.404$    & $3.101\pm 0.779$   & $0.405\pm 0.066$     & $0.340\pm 0.114$ \\
                      & 200         & $2.669\pm 0.118$     & $2.686\pm 0.294$  & $0.392\pm 0.013$     & $0.374\pm 0.020$  \\
                      & 2000             & $2.887\pm 0.045$     & $3.037\pm 0.517$  & $2.887\pm 0.045$     & $3.037\pm 0.517$  \\
\cline{1-6}
\multirow{3}{*}{2000} & 20           & $2.974\pm 0.626$    & $2.771\pm 0.751$  & $0.481\pm 0.090$     & $0.474\pm 0.096$    \\
                      & 200         & $2.710\pm 0.100$     & $2.628\pm 0.357$  & $0.415\pm 0.020$     & $0.412\pm 0.019$   \\
                      & 2000           & $2.850\pm 0.071$     & $2.913\pm 0.129$    & $0.423\pm 0.014$     & $0.420\pm 0.013$   \\                             
\hline
\end{tabular}}
\label{tab:use_case_2_div_values}
\end{table}

\bibliographystyle{ieeetr}
\bibliography{main}

\end{document}